\def\year{2018}\relax
\documentclass[letterpaper]{article} 
\usepackage{aaai18}  
\usepackage{times}  
\usepackage{helvet}  
\usepackage{courier}  
\usepackage{url}  
\usepackage{graphicx}  
\usepackage{amsmath}
\usepackage{amssymb}
\usepackage{bm}
\usepackage{color}
\usepackage{subcaption}

\begin{document}
%

\newcommand{\pos}{\ensuremath{\mathbf{p}}}
\newcommand{\vel}{\ensuremath{\mathbf{v}}}
\newcommand{\angVel}{\ensuremath{\bm{\omega}}}
\newcommand{\force}{\ensuremath{\mathbf{f}}}
\newcommand{\torque}{\ensuremath{\bm{\tau}}}
\newcommand{\forceConst}{\ensuremath{c}}
\newcommand{\torqueConst}{\ensuremath{\mathcal{I}}}
\newcommand{\gravity}{\ensuremath{\mathbf{g}}}
\newcommand{\stepsize}{\ensuremath{\Delta t}}
\newcommand{\step}{\ensuremath{q}}
\newcommand{\mR}{\mathbb{R}}

\newcommand{\note}[1]{}

\title{BubbleTouch: A Quasi-Static Tactile Skin Simulator
}
\author{Brayden Hollis, Stacy Patterson, Jinda Cui, Jeff Trinkle\\
Department of Computer Science\\
Rensselaer Polytechnic Institute, Troy, NY 12182, USA\\
bhollis13@my.whitworth.edu, sep@cs.rpi.edu, cuij2@rpi.edu, trink@cs.rpi.edu
\thanks{This work was partially supported by the National Science Foundation through the National Robotics Initiative (grant CCF-1537023).}
}
\maketitle
\vspace*{-1.2cm}
\begin{abstract}
We present BubbleTouch, an open source quasi-static simulator for robotic tactile skins. BubbleTouch can be used to simulate contact with a robot's tactile skin patches as it interacts with humans and objects.  The simulator creates detailed traces of contact forces that can be used in experiments in tactile contact activities. We summarize the design of BubbleTouch and highlight our recent work that uses BubbleTouch for experiments with tactile object recognition. 
\end{abstract}

\section{Introduction}

\noindent As robot technology advances, so does the vision of integrating and leveraging robots in unstructured human environments. For example, robots can provide in-home assistance and medical care for the elderly and people with disabilities~\cite{Cha2015}; robots can work side-by-side with human counterparts~\cite{nri}; and robots can work autonomously in environments that are unsafe for humans, for example, disaster areas.
To realize this vision, robots require improved tactile perception to detect and adapt to uncertain, dynamic physical environments and to interact in ways that are safe to themselves, and more importantly, to the humans around them. 
Further, robots need touch-based communication skills so that they can collaborate with people more efficiently and build good working relationships with them.

A key component of tactile perception and communication is \emph{tactile sensing}, defined as the continuous sensing of contact forces on a robot's surfaces~\cite{pennywitt1986robotic}.  Tactile sensing is provided by a \emph{tactile skin} that consists of a network or multiple networks of sensing elements, or \emph{taxels}, that measure quantities based on interaction with the environment, such as contact force and temperature. It is argued that, to safely interact with humans in uncontrolled environments, the tactile skin must cover the entire robot body~\cite{lumelsky2005sensing}, and to achieve human-like perception and concomitant dexterous manipulation skills, the skin should have spatial resolution as high as one taxel per square millimeter. This translates to on the order of tens of thousands to millions of individual taxels for human adult-sized robots, which have about 1.5 to 2 square meters surface area.\\
\vspace*{-0.cm}
Further, a data processing rate of 1kHz for high-fidelity force control is desirable~\cite{dahiya10tactile}.

Tactile skin data is critical to researchers studying physical human-robot interaction, however, tactile skin hardware is not readily available.  While some large-scale tactile systems are commercially available~\cite{Tekscan,PPS}, they are expensive and
difficult to mount on a robot in a way that covers most of its surfaces.  Tactile skin simulators provide a practical alternative for generating data, but few are available.  
The most notable is
SkinSim~\cite{habib2014skinsim}, which has been used to study to sensor fusion and feedback control during physical human-robot interaction.  A weakness of SkinSim is that simulations are prone to dynamic instability and numerical noise.  These problems arise from the dynamic contact model used by the Gazebo physics engine upon which SkinSim is built. 
 The instabilities are exacerbated when the interacting bodies have very large mass ratios, such as when a humanoid robot interacts with its environment. 

To support research on robotic tactile perception, we have been developing BubbleTouch~~\cite{bubbletouch},  an open-source quasi-static simulator for robotic tactile skins that avoids the  instabilities observed in SkinSim.
BubbleTouch is able to reliably simulate taxel skin patches of any size and shape, consisting of tens of thousands of taxels (to date), over arbitrarily fine timescales.   With BubbleTouch, researchers can design various contact trajectories, using different types of objects. The simulator creates detailed traces of contact forces that can be used in experiments. We summarize the design of BubbleTouch and highlight our recent work using BubbleTouch for tactile object recognition.

\begin{figure}
\centering
\includegraphics[width = .55\linewidth]{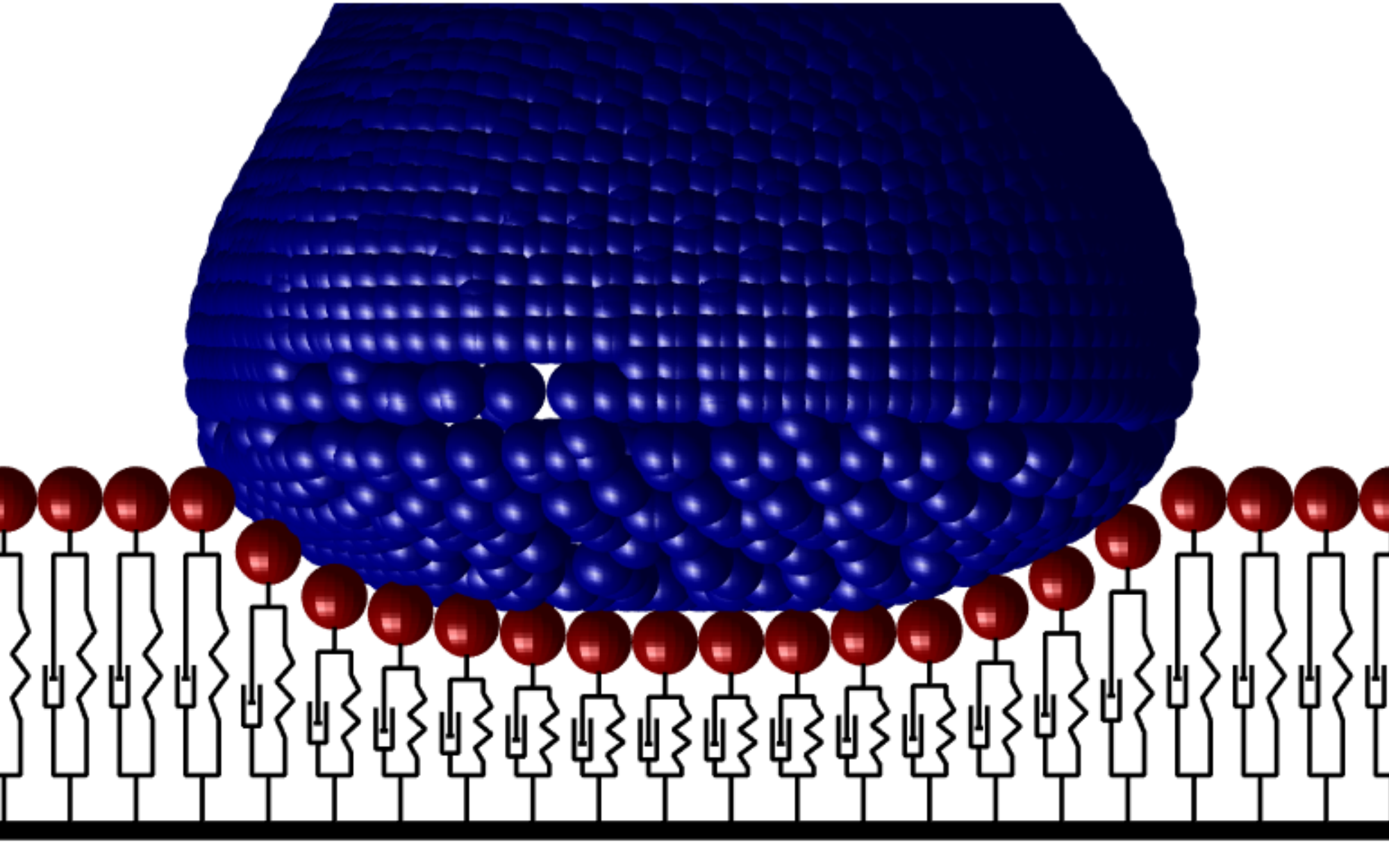}
\caption{Object in contact with tactile skin patch in BubbleTouch.}
\label{bubbletouchinteraction.fig}  
\end{figure}

\section{BubbleTouch Simulator Description}
BubbleTouch models four components that interact to generate tactile signals: taxels, robots, objects, and the world.

The tactile skin is composed of patches, each with a rigid substrate upon which many taxels are mounted.  
Each taxel is modeled as a sphere-spring-damper unit. 
Figure~\ref{bubbletouchinteraction.fig} show an edge view of a planar substrate to which red spheres are attached by springs and dampers.   Also shown is a portion of an object pressing into the skin, causing compression of springs and dampers, which translate to contact forces between the red and blue spheres.  

A robot is modeled as a rigid multi-link body covered by skin patches.  As the robot moves, the skin patches move with it.  Currently, the robot's motion is specified by imposing displacement trajectories for each joint.  This completely specifies the motions of the substrates of the skin patches.  Figure~\ref{bubbletouchinteraction.fig} is a snapshot from the robot moving into contact with the handle of a drill.

Objects in BubbleTouch are modeled as unions of rigid spheres. Modeling the touchable portion of each taxel as a sphere and the object as a union of spheres significantly speeds collision detection, which must be performed at every time step of the simulation.  Another benefit is that by adjusting the sphere radii, objects can be modeled as accurately as desired.  
Objects can be set as fixed and immovable, or objects can be specified to move according to a user-designed motion trajectories.

The world contains all robots and objects, and it models the gravity force vector.  The world model is also used to specify the simulation time-step size and simulation length.


Given the motions of objects and robots as inputs, 
 the simulator computes the deflections of the spheres and their contact forces.  This is done using a quasi-static contact model~\cite{PTLcoap96}, as opposed to a dynamic one.  As such, the taxels are in equilibrium at all times as they move, and therefore, the simulation does not become unstable.  In the current model, the taxels are restricted to move perpendicular to their substrate.  Since taxels cannot contact other taxels, the equations of motion (one for each taxel in contact with an object) are low-dimensional and uncoupled, which makes them particularly easy to solve.

\begin{figure}
\centering
\includegraphics[width = .5\linewidth]{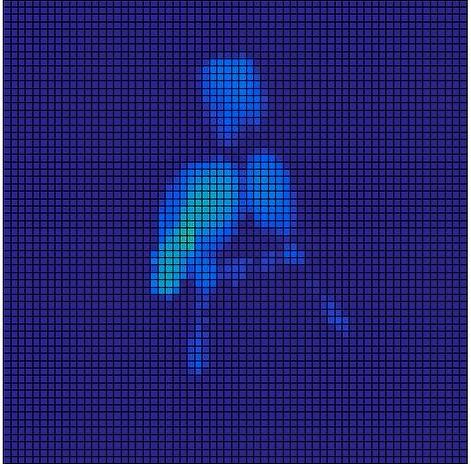}
\caption{Heat map of contact forces between a simulated human hand and a 64$\times$64-taxel array.}
\label{heatmap.fig}  
\end{figure}

The output of a BubbleTouch simulation is a trajectory of tactile contact forces. Each element of the trajectory is a vector containing the displacement of each taxel at a single time instance, and the trajectory is the sequence of these vectors. An example output for a single time instance is shown in Figure~\ref{heatmap.fig}, which shows contact between a (simulated) human hand and a 64$\times$64-taxel array.

In post-processing, we can translate the output displacement histories into forces.  For each taxel, the force is simply the sum of the spring force (its displacement times its stiffness constant) and the damper force (its displacement rate times its damping constant). We can also apply difference noise models that mimic physical hardware, for example, additive zero-mean Gaussian noise, which is consistent with the type of noise we have observed in individual tactile sensors in our lab's Barrett hand. Finally, we can apply Gaussian smoothing to simulate the effects of covering the skin with a layer of protective material.

\begin{figure}
 \centering
  \begin{subfigure}[t]{.23\linewidth}
\centering
\includegraphics[width = .6in]{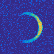}
\caption{\footnotesize Soup can.}\label{soup_g.fig}
\end{subfigure}
  \begin{subfigure}[t]{.26\linewidth}
\centering
\includegraphics[width = .6in]{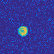}
\caption{\footnotesize Bottle.}\label{mustard_g.fig}
\end{subfigure}
  \begin{subfigure}[t]{.24\linewidth}
\centering
\includegraphics[width = .6in]{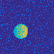}
\caption{\footnotesize Cleanser.}\label{bleach_g.fig}
\end{subfigure}
  \begin{subfigure}[t]{.23\linewidth}
\centering
\includegraphics[width = .6in]{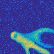}
\caption{\footnotesize Scissors.}\label{scissors_g.fig}
\end{subfigure}
\caption{Sample tactile signal snapshots of various household objects, with signals perturbed by Gaussian noise.}
\label{objects.fig}
\end{figure}

\section{Application: Tactile Object Recognition}
In recent work~\cite{Hollis2018}, we used BubbleTouch for research on tactile object recognition in high-resolution tactile skins. Using tactile object recognition, a robot can identify objects outside its line of site. This capability can aid in grasping occluded objects and in navigating in low visibility environments. 
We developed a framework that uses a compressed sensing-based data acquisition method to collect compressed versions of tactile signals, generated by the robot touching objects from above. Our framework uses a soft-margin support vector machine to classify the object based on the compressed tactile signal.

Using BubbleTouch, we were able to validate the efficacy of our framework in recognizing household objects based purely on single touches. Figure~\ref{objects.fig} 
shows examples of the uncompressed tactile signals generated by BubbleTouch that we used for our classification experiments. BubbleTouch enabled us to study the impacts of using arrays of different resolutions and sizes, as well as to easily generate data for thousands of contact experiments.

\vspace*{-1mm}
\section{Conclusion}
\vspace*{-1mm}
We have presented BubbleTouch, a tactile skin simulator.  BubbleTouch can reliably simulate contact interactions for robots equipped with large-scale tactile skins. 
We believe that such simulators are crucial in expanding research into touch-based human-robot interaction and communication.

We plan to extend BubbleTouch to include simulation of 3D force sensors. We also plan to expand the collection of object input plans to include human interactions that are used for communication, such as touches, taps, and shoves.

\bibliographystyle{aaai}
\bibliography{thesis}
\end{document}